\documentclass[letterpaper,10pt,conference]{ieeeconf}
\IEEEoverridecommandlockouts
\usepackage{flushend}
\usepackage{graphicx}
\usepackage{amsmath}
\usepackage{amssymb}
\usepackage{pgfplots}
\usepackage{grffile}
\usepackage{etoolbox}
\usepackage[space]{cite}
\usepackage[percent]{overpic}
\usepackage[font=footnotesize]{caption}
\usepackage{subcaption}
\usetikzlibrary{plotmarks}
\pgfplotsset{compat=newest}
\pgfplotsset{plot coordinates/math parser=false}
\AtEndDocument{\par\leavevmode}

\newlength\figurewidth
\newlength\figureheight
\newcommand{\dif}{\mathrm{d}}
\newcommand{\dir}{\omega}
\newcommand{\hem}{\mathcal{H}^2}
\newcommand{\pt}{\mathrm{p}}

\title{\LARGE\bf Light Source Estimation with Analytical Path-tracing}
\author{Mike Kasper, Nima Keivan, Gabe Sibley, Christoffer Heckman$^\ast$
\thanks{This work was graciously supported by Toyota Motor Corporation.}
\thanks{All listed authors are with the Autonomous Robotics and Perception Group
at the University of Colorado, Boulder.}
\thanks{$^\ast$Corresponding author. E-mail: {\tt\small christoffer.heckman at
colorado.edu}}}
\begin{document}
\maketitle
\thispagestyle{empty}
\pagestyle{empty}


\begin{abstract}


We present a novel algorithm for light source estimation in scenes reconstructed
with a RGB-D camera based on an analytically-derived formulation of
path-tracing. Our algorithm traces the reconstructed scene with a custom
path-tracer and computes the analytical derivatives of the light transport
equation from principles in optics. These derivatives are then used to perform
gradient descent, minimizing the photometric error between one or more captured
reference images and renders of our current lighting estimation using an
environment map parameterization for light sources. We show that our approach of
modeling all light sources as points at infinity approximates lights located
near the scene with surprising accuracy. Due to the analytical formulation of
derivatives, optimization to the solution is considerably accelerated. We verify
our algorithm using both real and synthetic data.

\end{abstract}


\section{Introduction}
\label{sec:introduction}


Computer vision is often referred to as the ``inverse graphics'' problem. This
is because many of the equations and relations used in computer vision find
their roots in the understanding of image formation and light interaction.
However the complete process of image formation is mainly ignored in most
applications of computer vision. In the case of localization and mapping,
brightness constancy is assumed from multiple viewpoints and robust estimation
is used to reduce or ignore the influence of any non-cooperative observations.
While this approach works well for surfaces exhibiting Lambertian reflectivity,
specular and transparent surfaces are often treated as outliers. This
assumptions is also the cause of the complexity of localization from a
completely dense 3D map. While brightness constancy may apply from small
baselines, it generally does not in the case of widely varying baselines.

These approximations inhibit the estimation of many quantities of interest,
such as light position, sensor response curves, and surface properties of
in-scene objects. The light position estimation problem itself has garnered
significant interest in the autonomous robotics community. Dynamic shadow
effects for instance stymie feature- and deep learning-based algorithms for
place and object recognition. \cite{IROS_2013_Corke} proposed a filter-based
approach to this problem which removes shadows from images, but struggles with
shadows containing reflected light and artificial light sources, such as lamps
or headlights. The assumption of constant illumination is critical to
tracking applications \cite{IJRR_2016_Whelan}, which could be potentially
improved through the consideration of irregular illumination imbued by
individual light sources as well.

A number of previous works have handled in-scene light source position
estimation. For instance, Elastic Fusion \cite{IJRR_2016_Whelan} introduces a
method for this via ray casting and tessellating the space with potential light
source voxels with a merging function. This method precludes the estimation of
surface reflectancies however, as ray casting requires that they are provided
at the outset. As an alternative to using potential light emitting voxels, an
``environment map'' may instead be employed to represent light sources. This
approach imposes an encapsulating 2-manifold (typically a hemisphere)
around the 3-dimensional region of interest and parametrizes a coordinate
system on the manifold that act as inwardly-directed point light sources
\cite{ISMAR_2012_Jachnik, 3DV_2014_Lalonde}.

A number of approaches have been taken on estimating the effect of in-scene
lighting with considerable success. For instance, \cite{ISMAR_2013_Meilland}
addresses light source estimation by laying out a light-field which can then
be used for casting shadows. The light sources in this case do not follow a
physically derived model, but recreate in-scene lighting with impressive
accuracy. A different approach is applied by \cite{ISMAR_2014_Knorr}, where
a learned set of radiance transfer functions are applied for generating
plausible lighting incident on a human face. Meanwhile, there has been steady
development in approaches to estimating light effects for out-of-scene sources,
e.g.\ \cite{ECCV_2002_Zhou, ECVM_2007_Takai, ICCV_2005_Hara, BMVC_2013_Boom}.
These approaches may be used to e.g.\ cast artificial shadows for the purposes
of augmented reality, but do not admit the arbitrary estimation of in-scene
parameters.


\begin{figure}
  \centering
  \includegraphics[width=0.239\textwidth]{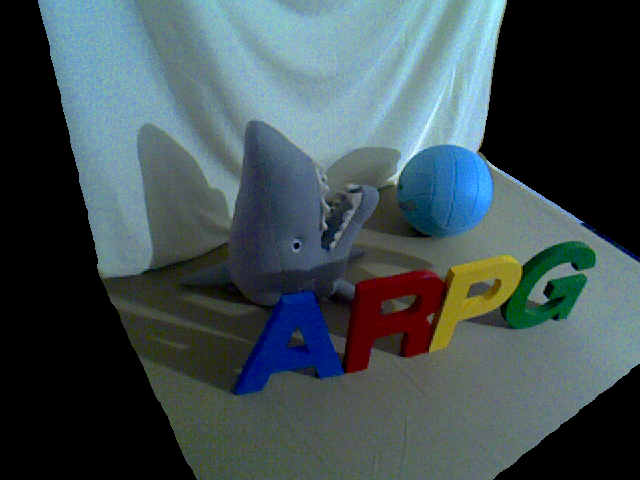}
  \includegraphics[width=0.239\textwidth]{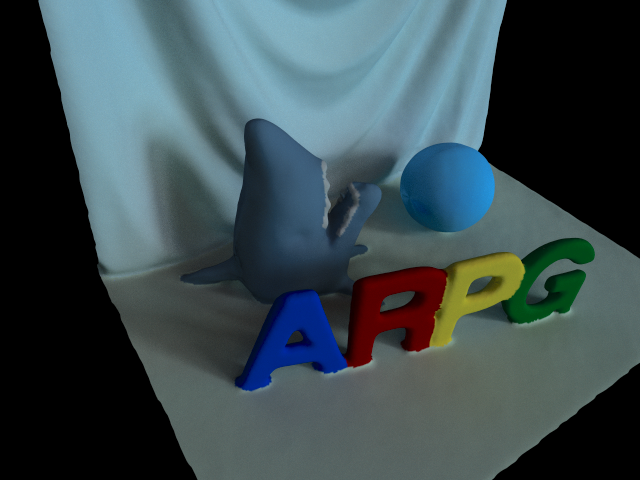}
  \caption{\emph{Left}: captured reference image, \emph{Right}: render of
  estimated lighting conditions found using our algorithm}
  \label{fig:intro}
\end{figure}


To address these limitations, optical path tracing may be employed to both
estimate scene parameters as well as for photorealistic rendering, including
the effects of lens flares, specular highlights, shadows, and other visual
phenomena. This approach applies a generative model for light interaction with
a scene using methods from optics, and admits the selective approximation as
need warrants. Path tracing however has proven to be a prohibitively
computationally expensive solution, as path tracing is a nonlinear operation
with a thousands of optimization parameters.

The method described in this paper eschews the use of finite differences in
favor of analytical derivatives of the light transport equation. We use an
environment map parametrization of light sources and employ robust nonlinear
optimization in order to estimate the position of light sources in a scene.
Our process is naturally extensible to the estimation of scene properties
including surface reflectivity. We give a description of our method in Sections
\ref{sec:overview}-\ref{sec:light_source_estimation}. Our results are provided
in Section \ref{sec:experimental_results} and discussed in Section
\ref{sec:discussion}. We draw conclusions from our work in Section
\ref{sec:conclusion}.


\section{Overview}
\label{sec:overview}

\begin{figure}
  \centering
  \includegraphics[width=0.48\textwidth]{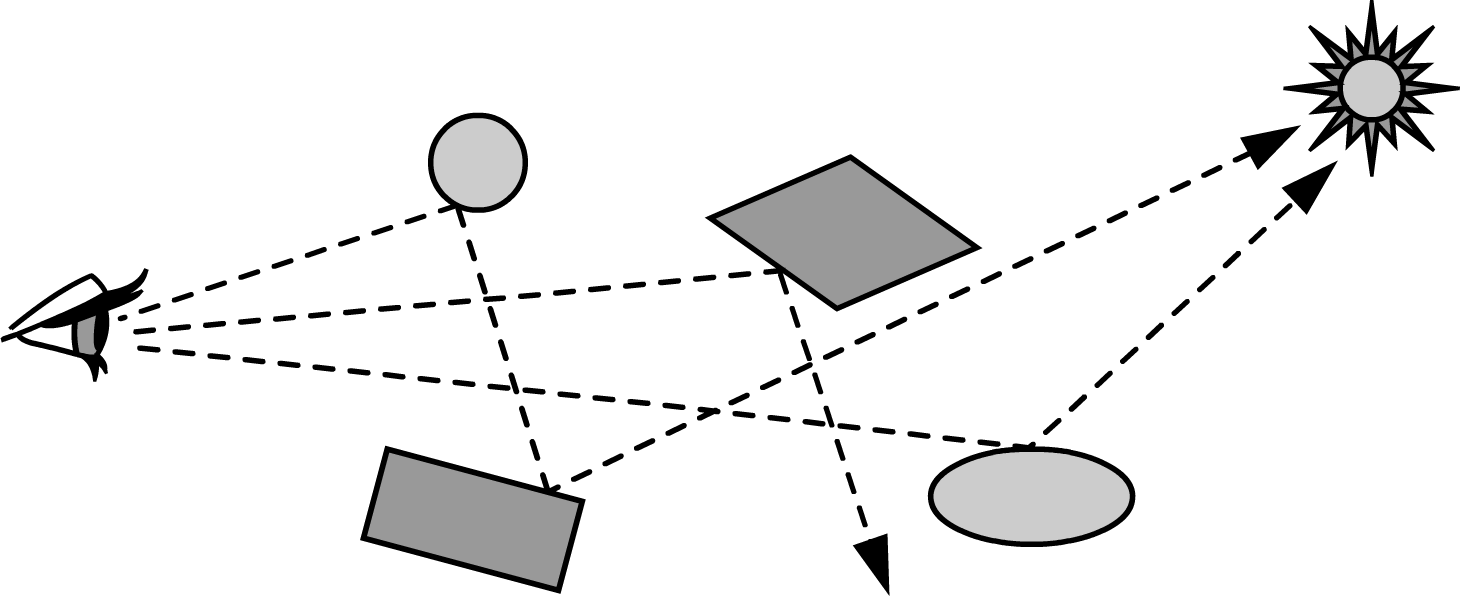}
  \caption{Path-tracing models how rays of light traverse a given scene. With
  each collision with the scene's geometry, information about surface materials,
  angle of incidence, and location of the light source are used to compute the
  amount of light reaching the synthetic camera sensor.}
  \label{fig:path_tracing}
\end{figure}

Physically-based rendering refers to the process of image formation that remains
as faithful to real-world optics as possible. This is achieved through
accurately modeling the interaction of light with the surfaces and volumes in
our scene. Such rendering engines are often referred to as path-tracers, as they
simulate the path an emitted ray of light takes as it traverses the scene
before eventually being captured by our synthetic image sensor. As the vast
majority of light in the scene will never actually reach the image sensor, this
problem is typically inverted for the sake of computational simplicity. So we
instead trace rays of light from the image sensor back to the light source. In
order to synthesize high-fidelity images with little to no noise, hundreds or
thousands of rays must be traced for every pixel. These general concepts are
illustrated in Figure \ref{fig:path_tracing}.

The most important concept to take away from all of this, in regards to the work
presented in this paper, is that while this process requires tracing thousands
of rays, intersecting them with the scene geometry at each bounce, computing the
material properties and angles of incidence at each point of intersection, as
the rays make their way towards a light source, all of the geometry, material
properties, and math compute \emph{a single coefficient} which denotes the
amount of light from a given light source that is stored in a given pixel. This
idea sits at the heart of our algorithm.

We leverage this fact in our formulation of the light estimation problem. A
custom path-tracer generates these coefficients for each pixel-light pair. We
use these values to compose a linear system of equations that can be solved
using standard optimization techniques. The power of this approach is that not
only can it be gracefully extended to handle any optical phenomema that could
be present in our observed scene, it can also be rewritten to solve for
different components of image formation.


\section{Scene Model}
\label{sec:scene_model}

We now describe the input required by our algorithm for performing path-tracing
and estimating the lighting conditions in the observed scene.


\subsection{3D Geometry}
\label{sec:geometry}

In order to properly simulate how rays emitted from light sources bounce off
surfaces before reaching our camera we first require a 3D geometric
representation of the scene. Our implementation uses a triangle mesh with
corresponding surface normals as seen in Figure \ref{fig:mesh}. We capture such
a mesh using a Asus Xtion Pro Live and the InfiniTam 3D reconstruction
framework \cite{TVCG_2015_Kahler}. The more complete this reconstruction is the
better we can recreate shadows and simulate how light bounces around the
surfaces of the scene. However we have observed that even largely incomplete
reconstructions still afford accurate light source estimation.


\subsection{Surface Albedos}
\label{sec:surface_albedos}


To render a RGB image of our current lighting estimation we must have an albedo
associated with each vertex in our mesh. The albedo describes the underlying
color of an object at a given point, void of shadows or any other shading
information. Figure \ref{fig:albedo} illustrates this concept. The problem of
separating albedos and shading information found in images, often referred to as
\emph{intrinsic image decomposition}, is the subject of a rich field of ongoing
research \cite{ICCV_2013_Chen, ICCV_2015_Hachama, TG_2015_Duchene}. Given a
decomposed albedo image, we can map the albedos onto the surface of the mesh
using existing color mapping techniques \cite{ACM_2014_Zhou,
IJCV_2014_Goldlucke}. However, to obviate this challenge we currently assume
albedo associations are known, although this knowledge need not be perfectly
accurate.


\subsection{Reference Images}
\label{sec:reference_images}

\begin{figure}
  \centering
  \begin{subfigure}{0.1575\textwidth}
    \includegraphics[width=\textwidth]{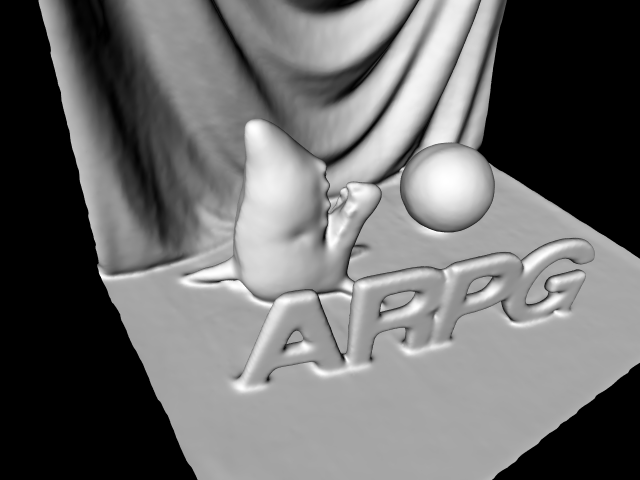}
    \caption{}
    \label{fig:mesh}
  \end{subfigure}
  \hfill
  \begin{subfigure}{0.1575\textwidth}
    \includegraphics[width=\textwidth]{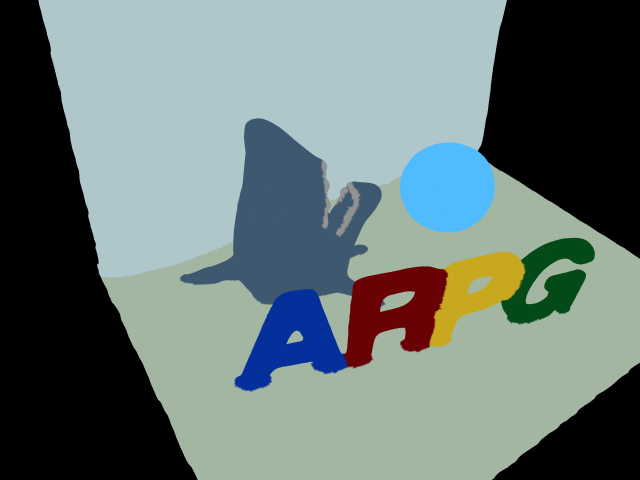}
    \caption{}
    \label{fig:albedo}
  \end{subfigure}
  \hfill
  \begin{subfigure}{0.1575\textwidth}
    \includegraphics[width=\textwidth]{figures/intro_actual.png}
    \caption{}
    \label{fig:reference}
  \end{subfigure}
  \caption{Input to our algorithm consists of (\subref{fig:mesh}) 3D geometry,
  (\subref{fig:albedo}) surface albedos, and (\subref{fig:reference}) one or
  more masked reference images with corresponding camera intrinsics and 6-DOF
  pose.}
\end{figure}

Our optimization works to minimize the photometric error between renders of our
current lighting estimation and one or more reference images. To avoid
penalizing differences resulting from an incomplete geometric reconstruction, we
mask the reference images such that pixels corresponding to holes in our mesh
are ignored, as seen in Figure \ref{fig:reference}. In order to render
synthetic images that can be compared directly with the captured reference
images we must know the 6-DOF pose and intrinsics of the camera used. We
calibrate the Xtion Pro Live camera using the Calibu calibration framework
\cite{APRG_2016_Calibu}. This calibration provides the camera intrinsics
matrix, distortion parameters, and the IR-to-RGB camera transform.

The pose of the IR camera is estimated as InfiniTam reconstructs the scene
geometry \cite{TVCG_2015_Kahler}. We then compute the pose of the RGB camera
using the IR-to-RGB camera transform previously estimated. As we are directly
comparing our synthetic images with the captured reference images, it is also
critical that we account for the camera's response curve and vignetting. We
estimate the camera's response curve functions using \cite{CVPR_1999_Mitsunaga}.
These response curves are then used to convert captured image intensities to
irradiance values. We then account for the camera's vignetting using
\cite{TPAMI_2008_Kim}. While multiple reference images can be used in our
optimization we find a single well-placed image is often sufficient to
accurately estimate the lighting conditions.


\subsection{Environment Light}
\label{sec:environment_light}


In this work we model light using an environment map \cite{ISMAR_2012_Jachnik,
3DV_2014_Lalonde, ISMAR_2014_Rohmer}. Instead of sampling points in 3D space,
environment map lighting admits directional sampling. This representation works
best when approximating lights located further from the observed scene. While
many works have considered in-scene lighting examples \cite{ISMAR_2013_Meilland,
ISMAR_2014_Knorr}, we instead focus on out-of-scene sources
\cite{ECCV_2002_Zhou, ECVM_2007_Takai, ICCV_2005_Hara, BMVC_2013_Boom}. To
compute the direct incident radiance from our environment map $L_d$ arriving at
a point $\pt$ we trace a ray with origin $\pt$ in some direction $\dir$. If the
ray is unobstructed by the scene geometry, point $\pt$ will receive the full
radiance traveling along $\dir$ as determined by the environment map.

To compute the radiance emitted by the environment map along a given direction,
we first discretize a unit sphere into a finite number of uniformly spaced
points, with each point representing a direction that can be sampled. We perform
a similar discretization as described in \cite{ISMAR_2012_Jachnik} over an
entire sphere. The resolution of this discretization is indicated by the number
of desired rings. The spacing of points around each ring is computed to be as
close to the inter-ring spacing as possible, as seen in Figure
\ref{fig:environment_light}. When tracing a ray along a given direction we
determine the nearest-neighbor direction from the discretized environment map
and return its associated RGB value $\lambda$ as the emitted radiance.


\section{Light Transport Equation}
\label{sec:light_transport_equation}



In this section we provide a brief overview of the \emph{light transport
equation} (LTE), so that the reader may better understand how we perform our
light source estimation. Intuitively, the LTE describes how radiance emitted
from a light source is distributed throughout a scene. Formally, we compute the
exitant radiance $L_o$ leaving a point $\pt$ in direction $\dir_o$ as:


\begin{figure}
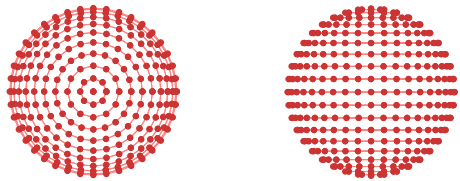

  \centering
  \setlength\figurewidth{0.15\textwidth}
  \setlength\figureheight{0.15\textwidth}
  \input{figures/sphere_top.tex}
  \hspace{15pt}
  \setlength\figurewidth{0.15\textwidth}
  \setlength\figureheight{0.15\textwidth}
  \input{figures/sphere_side.tex}
  \caption{\emph{Left}: Top-down and \emph{Right}: side view of our environment
  map discretization. The light depicted here consists of 21 rings and 522 total
  discrete directions.}
  \label{fig:environment_light}
\end{figure}


\begin{equation}
\label{eq:lte_integral}
L_o(\pt, \dir_o) = \int_{\hem} f(\pt, \dir_o, \dir_i) L_i(\pt, \dir_i)
    | \cos \theta_i | \dif\dir_i
\end{equation}

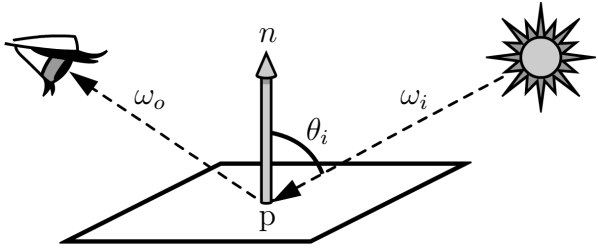
\begin{figure}
  \centering
  \begin{overpic}[height=0.18\textwidth]{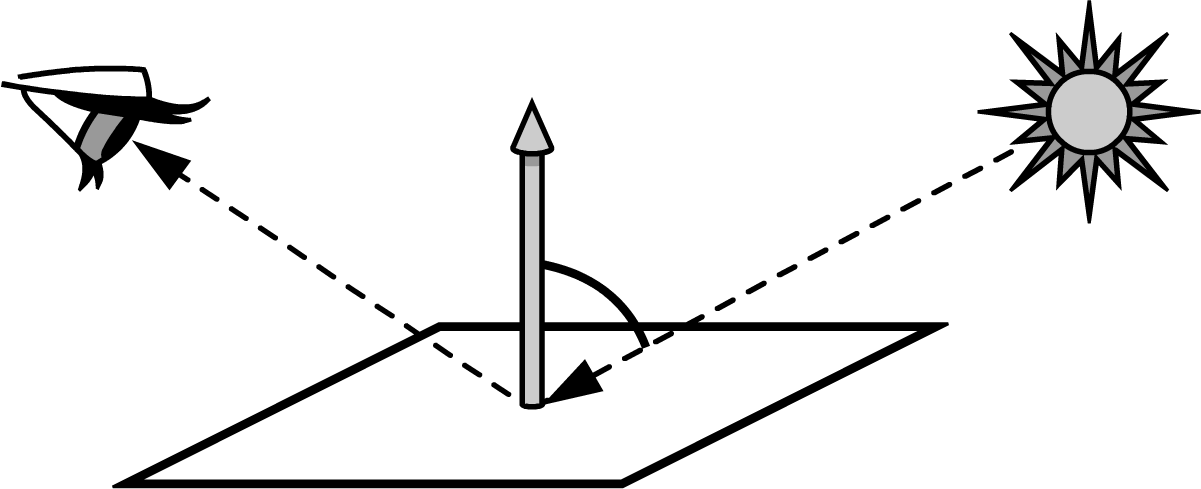}
   \put(43.0, 34.0){\large$n$}
   \put(43.0,  3.0){\large$\pt$}
   \put(22.0, 23.5){\large$\dir_o$}
   \put(67.0, 23.5){\large$\dir_i$}
   \put(51.0, 17.5){\large$\theta_i$}
  \end{overpic}
  \caption{Components of a single ray bounce in the LTE. $\pt$ is the point of
  intersection, $\dir_o$ is the direction of exitant radiance $L_o$, $\dir_i$ is
  the direction of incident radiance $L_i$, $n$ is the surface normal found at
  $\pt$, and $\theta_i$ is the angle between $n$ and $\dir_i$.}
  \label{fig:bounce}
\end{figure}

\noindent This integral evaluates the amount of incident radiance $L_i$ arriving
at $\pt$ over the unit hemisphere $\hem$ oriented with the surface normal found
at $\pt$. The function $f$ denotes the \emph{bidirectional reflectance
distribution function} (BRDF) found at point $\pt$. The BRDF defines how much of
the incident radiance arriving at $\pt$ along $\dir_i$ is reflected in direction
$\dir_o$. Finally, $\theta_i$ is the angle between $\dir_i$ and the surface
normal found at $\pt$. Figure \ref{fig:bounce} illustrates these
relationships. Using Monte Carlo integration we can rewrite Eq.\
\eqref{eq:lte_integral} as the finite sum:

\begin{equation}
\label{eq:lte_monte_carlo}
L_o(\pt, \dir_o) = \frac{1}{N} \sum_{i=1}^{N} \frac{f(\pt, \dir_o, \dir_i)
    L_i(\pt, \dir_i) | \cos \theta_i |}{p(\dir_i)}
\end{equation}

\noindent where $N$ is the number of directions $\dir_i$ sampled from the
distribution described by the \emph{probability density function} (PDF) $p$. The
incident radiance $L_i$ represents both the radiance coming directly from our
light source, which we denote by $L_d$, and the radiance reflected off
surrounding surfaces, which we denote by $L_r$. Given the recursive nature of
path-tracing, we can effectively rewrite the incident radiance $L_r$ as exitant
radiance $L_o$:

\begin{equation}
\label{eq:reflected_radiance}
L_r(\pt, \dir) = L_o(\pt', \dir)
\end{equation}

\noindent where $\pt'$ is the point where a ray leaving from $\pt$ in direction
$\dir$ first intersects the scene. As we recursively bounce rays around our
scene the BRDF, PDF and $\cos \theta$ terms of the LTE are compounded. We refer
to this product as \emph{throughput}. Formally, the throughput $T$ of the
$i^\text{th}$ point $p_i$ in our current path is defined as:

\begin{equation}
\label{eq:throughput}
T(\pt_i) = \prod_{j=1}^{i} \frac{f(\pt_j, \dir_{j-1}, \dir_j) | \cos \theta_j |}
    {p(\dir_j)}
\end{equation}


\noindent
With this definition of throughput, we can now define $L_d$ as:

\begin{equation}
\label{eq:direct_radiance}
L_d(\pt, \dir_i) = V(\pt, \dir_i) T(\pt) \lambda_i
\end{equation}


\noindent where $\lambda_i$ is the estimated radiance for the direction in our
environment map that is closest to $\dir_i$. The visibility function $V$
evaluates to 1 if a ray leaving from point $\pt$ in direction $\dir_i$ is not
obstructed by the scene geometry, and 0 otherwise.


To compute the final pixel intensity $I$, we integrate the intensities of all
rays $M$ arriving at the corresponding point on our synthetic
sensor (of the form of Eq.\ \eqref{eq:lte_monte_carlo}). Using Monte Carlo
integration we can evaluate this with the finite sum:

\begin{equation}
\label{pixel_mc}
I = \frac{1}{M} \sum_{i=1}^{M} \frac{L_o(\pt_i, \dir_i)}{p(\dir_i)}
\end{equation}


\noindent where $\pt_i$ refers to the point where a ray originating from our
sensor and traveling along $\dir_i$ first intersects with the scene. For more
information on path-tracing and the LTE see \cite{BOOK_2010_Pharr}.



\section{Light Source Estimation}
\label{sec:light_source_estimation}

In this section we present our algorithm for light source estimation. We
construct the optimization problem by first initializing all lighting
parameters to be near zero, resulting in a completely dark scene. As light
parameters are uniformly initialized, each light starts with the same
probability of being sampled.

\subsection{Objective}
\label{sec:objective}

Our objective function minimizes the photometric error between the set of
original reference image $\{I_o\}$ and a set of corresponding reconstructed
images $\{I_r\}$, which depict our current lighting estimation. So the
photometric error $E_p$ can be defined as:

\begin{equation}
E_p = \sum_i \sum_{\mathbf{p} \in P_i} (I_{o,i}(\mathbf{p}) -
    I_{r,i}(\mathbf{p}))^2
\end{equation}

Given the prevalence of large discrepancies caused by an imperfect scene
reconstruction or under-sampling, we employ the Cauchy robust norm on the raw
photometric error values. Due to the potential over-parameterization of
lighting, an activation penalty function is used as an additional cost. Each
light direction $\lambda_i$ has an associated activation penalty $\Phi_i$
defined as:

\begin{equation}
\label{eq:activation_cost}
\Phi_{i} = \alpha \log(1 + \beta \sum_{c=1}^n \lambda_{i, c})
\end{equation}

\noindent where $\alpha$ and $\beta$ are constant weights and $n$ is the number
of channels in our images. This logarithmic cost penalizes the initial
activation of a light direction but then plateaus as the intensity of light
increases. Increasing the weight of $\alpha$ and $\beta$ favor solutions with
fewer activated lights. The use of an activation cost has the additional benefit
of completely disabling lights that can never be sampled due to the scene
geometry and viewing angle of the reference images. While all lighting
parameters are initialized to be near zero, if the collective probability of
sampling unreachable lights is high enough, it could have a negative impact on
the variance of our path-traces. This concept is explained further in the follow
sections. With this additional term our final objectivation function $E$
becomes:

\begin{equation}
\label{eq:objective}
E = E_p + \Phi
\end{equation}

\subsection{Light Transport Derivatives}
\label{sec:light_transport_derivatives}


In order to perform our light optimization we first need to compute the Jacobian
of partial derivatives of pixel values with respect to the lighting parameters.
We initialize the Jacobian with all values set to zero. We then populate
the values of the Jacobian by tracing the given scene with our custom
path-tracer. For each pixel sample we construct the path a ray of light takes as
it bounces around the scene, while computing the throughput of the path as
described in Section \ref{sec:light_transport_equation}. When a ray eventually
samples the light source, we add the ray's final throughput to the corresponding
value in the Jacobian. Following Eq.\ \eqref{eq:direct_radiance}, we can
formally define the partial derivative of the pixel $I_i$ with respect to the
sampled light $\lambda_j$ for each color channel $c$ as the finite sum:

\begin{equation}
\label{eq:lte_deriv}
\frac{\partial I_{i,c}}{\partial \lambda_{j,c}} =
    \frac{1}{MN} \sum_{k=1}^K \frac{V(\pt_k, \dir_j) T_c(\pt_k)} {p(\dir_i)}
\end{equation}

\noindent where $K$ is the number of samples for pixel $I_i$ that sample
light source $\lambda_j$, $\pt_k$ is the final point on in path $k$ before
sampling the light source, and $T_c$ is the throughput value for color channel
$c$.


In additional to the partial derivatives corresponding to the images
formation, we also need to compute the partial derivatives for our activation
cost. Following Eq.\ \eqref{eq:activation_cost} we compute the partial derivative
of the activation cost $\Phi_i$ with respect to the light $\lambda_i$ for each
color channel $c$ as:

\begin{equation}
\label{eq:activation_derivative}
\frac{\partial \Phi_{i}}{\partial \lambda_{i, c}} =
    \frac{\alpha \beta} {1 + \beta (\sum_{j=1}^{n} \lambda_{i, j})}
\end{equation}


\subsection{Gradient Descent}
\label{sec:gradient_descent}

Having estimated our Jacobian, we can now perform the light location and
intensity optimization.  As our Jacobian will typically be extremely large,
dense, and exhibit no anticipated structure we can leverage, we employ gradient
descent with backtracking as inverting or decomposing the Jacobian would be
computationally prohibitive. To account for the inherent constraint that light
parameters cannot be negative we apply gradient projection after each update
\cite{JSIAM_1960_Rosen}.  We continue this process until the optimization has
converged, as indicated by the Wolfe conditions on gradient magnitude
\cite{BOOK_2006_Nocedal}.

\subsection{Sequential Monte Carlo}
\label{sec:sequential_monte_carlo}

As described in Section \ref{sec:light_transport_derivatives}, we leverage
importance sampling when tracing the scene, so that our estimated derivatives
will exhibit a lower amount of variance while using a smaller number of
samples. However, when we first construct our light optimization problem, we
initialize our lighting parameters uniformly. Consequently, all lights will
initially have the same probability of being sampled and we will not receive any
benefit from importance sampling. If we were to formulate the optimization
problem with a higher environment map resolution or use a smaller the number of
samples per pixel the variance in the estimated derivatives will increase,
resulting in a poorer estimate of the lighting parameters.

Yet once gradient descent converges and yields an updated set of lighting
parameters we can repeat the process of estimating the derivatives. We would no
longer be sampling lights uniformly, and we would now benefit greatly from
importance sampling. This technique is referred to as particle filtering or
sequential Monte Carlo (sMC) \cite{SORMS_2014_Homem}. Our algorithm continues to
re-estimate the Jacobian, retracing the scene and sampling lights according to
our latest estimate of lighting parameters, and performs gradient descent until
we no longer observe any significant change in the lighting parameters.


\section{Experimental Results}
\label{sec:experimental_results}

We verified our algorithm on several distinct real and synthetic datasets. All
lighting optimizations were performed using a environment map resolution of
between 9 to 21 light rings and a single reference image with a resolution of
$160\times120$. We first reconstructed scene geometries for two real scenes.
Albedos were manually associated with each vertices in the resulting mesh.
These scene models were then used with both real and synthetic reference images.

\begin{figure*}
  \centering
  \setlength{\tabcolsep}{2pt}
  \begin{tabular}{ccc}

  \includegraphics[width=0.28\textwidth]{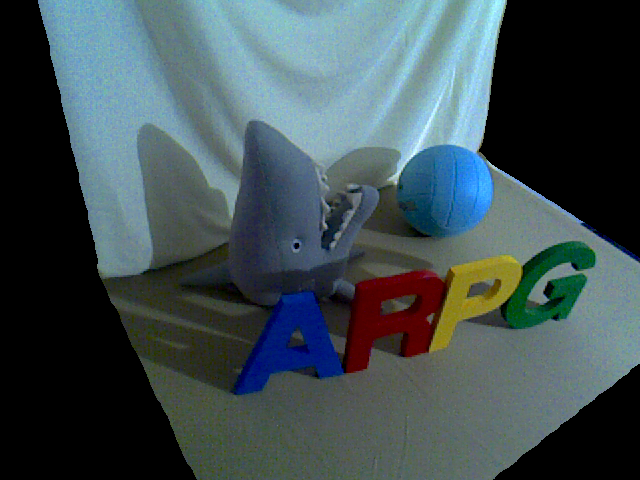}   &
  \includegraphics[width=0.28\textwidth]{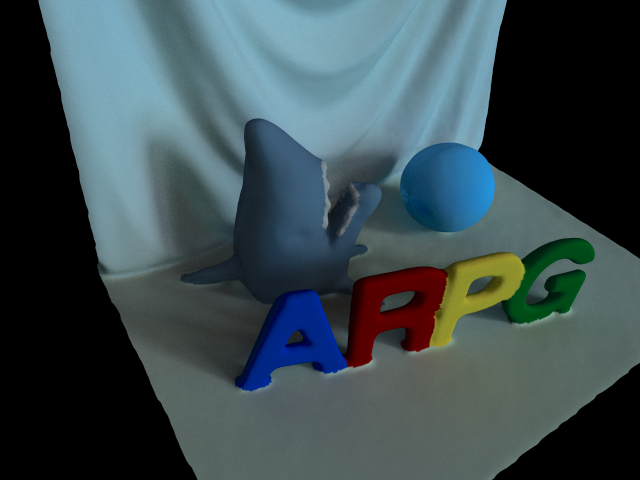} &
  \includegraphics[width=0.28\textwidth]{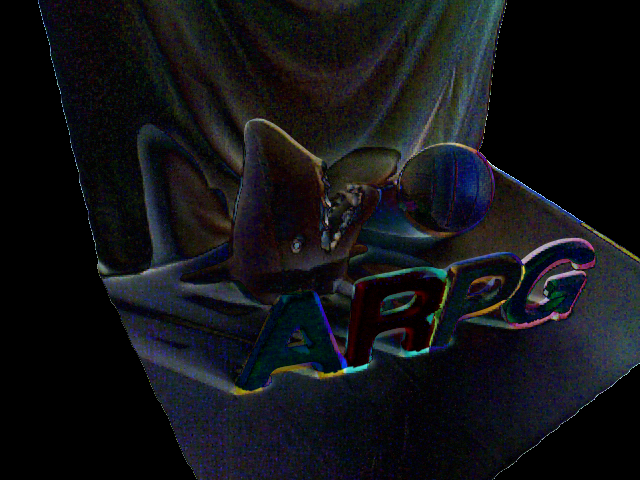}    \\

  \includegraphics[width=0.28\textwidth]{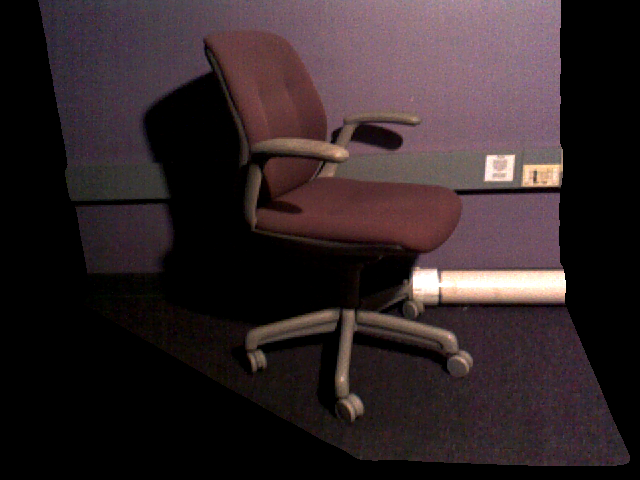}   &
  \includegraphics[width=0.28\textwidth]{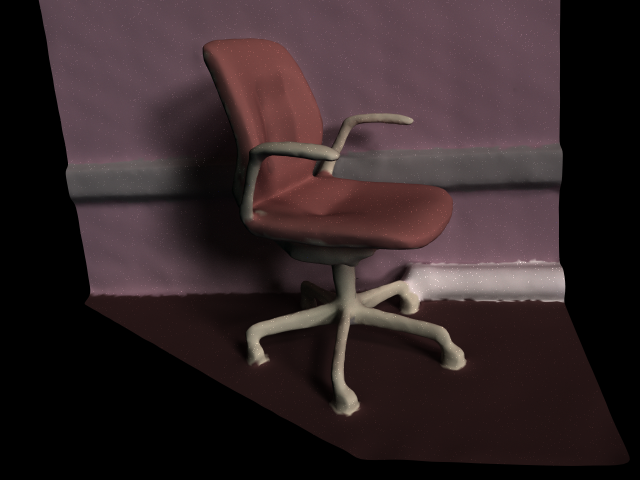} &
  \includegraphics[width=0.28\textwidth]{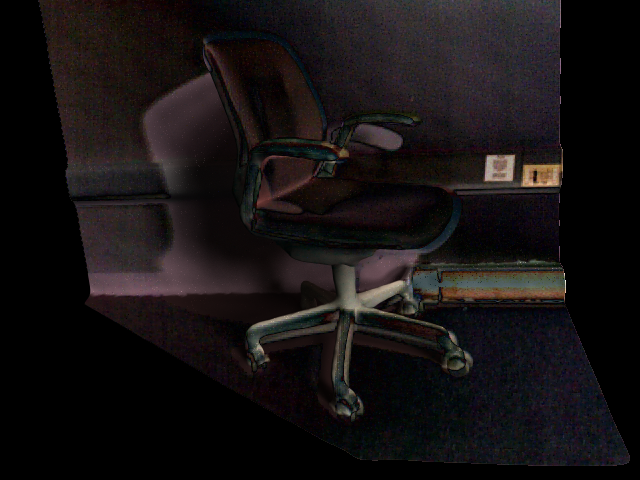}    \\

  \includegraphics[width=0.28\textwidth]{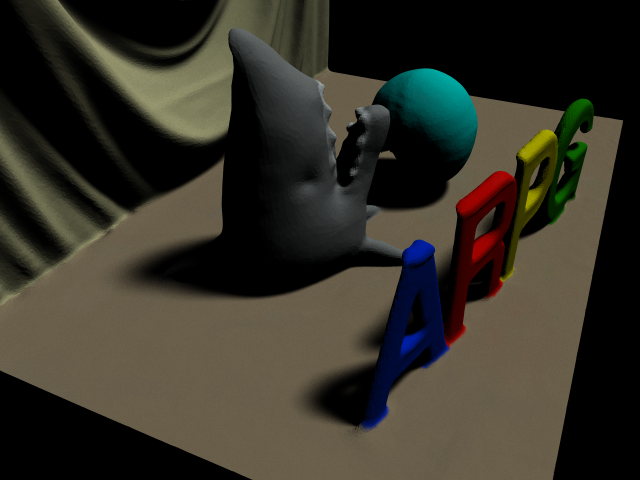}   &
  \includegraphics[width=0.28\textwidth]{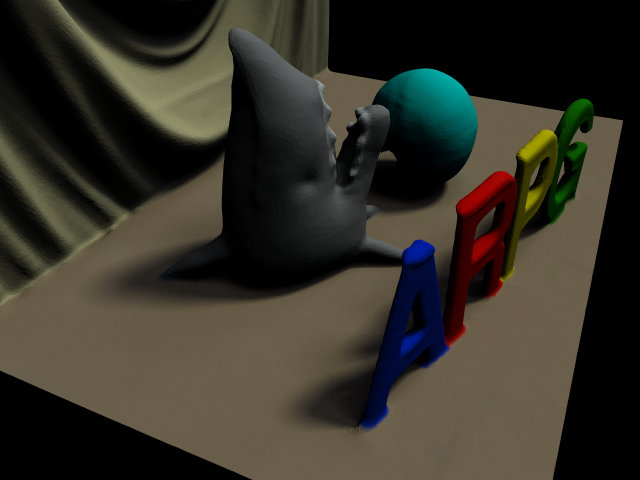} &
  \includegraphics[width=0.28\textwidth]{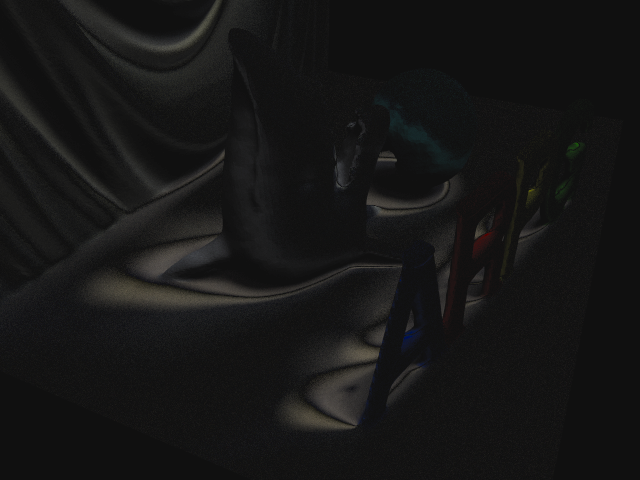}    \\

  \includegraphics[width=0.28\textwidth]{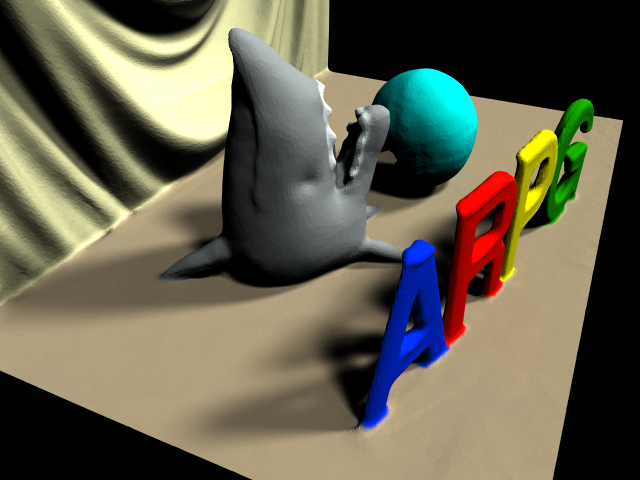}   &
  \includegraphics[width=0.28\textwidth]{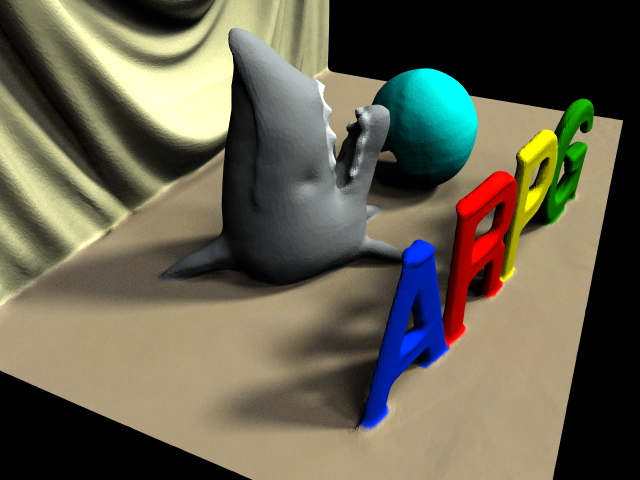} &
  \includegraphics[width=0.28\textwidth]{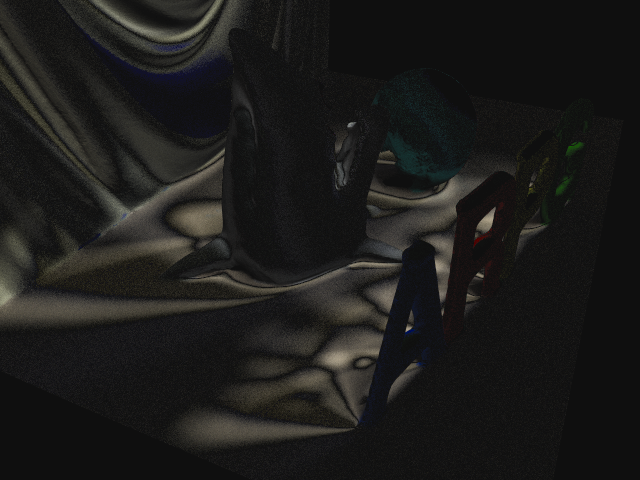}    \\

  \includegraphics[width=0.28\textwidth]{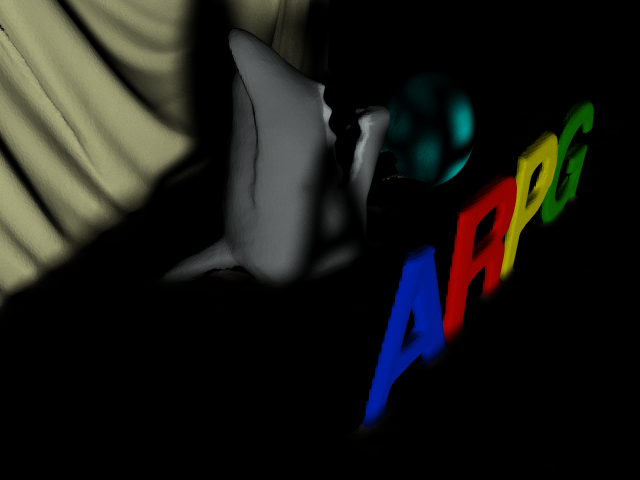}   &
  \includegraphics[width=0.28\textwidth]{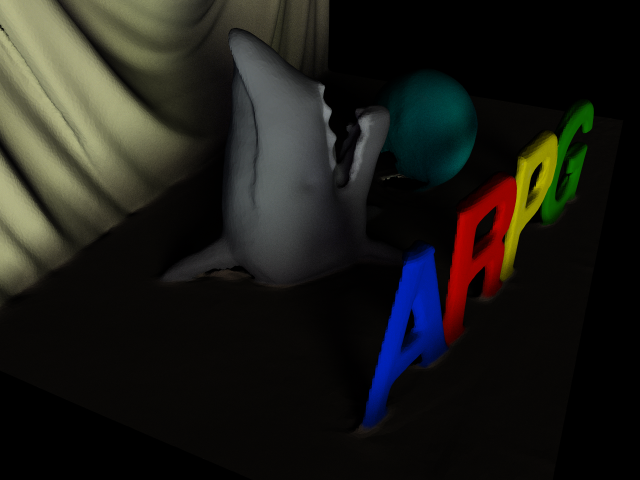} &
  \includegraphics[width=0.28\textwidth]{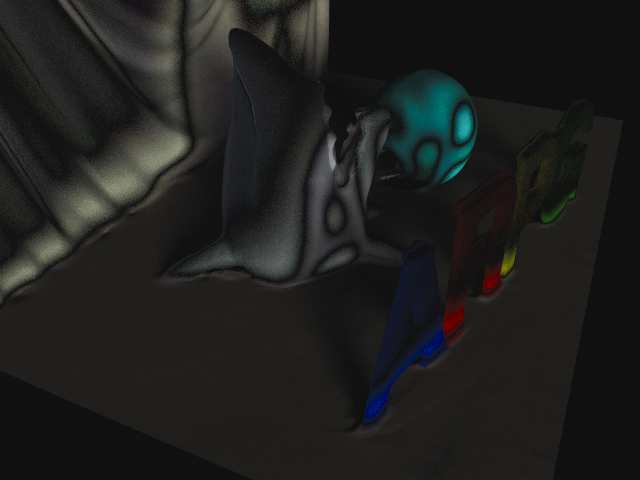}    \\

  \end{tabular}
  \caption{\emph{Left}: captured reference image, \emph{Middle}: render of
  estimated lighting conditions, \emph{Right}: photometric error. Rows 1 and 2
  use reference images captured with an Xtion Pro Live RGB-D camera. Rows 3-5
  use reference images rendered with our path-tracer using one or two area
  lights.}
  \label{fig:results}
\end{figure*}

The two experiments using real reference images were captured using Xtion Pro
Live while performing scene reconstruction. These scenes were illuminated with a
single white LED 1,100 lumen light bulb placed within 3 meters of the observe
scene. Results for these two experiments can be seen in rows 1 and 2 of Figure
\ref{fig:results}.

We performed three additional experiments with synthetically generated reference
images. There images were rendered using the same mesh and albedos used during
optimization. However, there were illuminated using one or more spherical area.
This was an intentional decision as not to use the same lighting model we are
trying to simulate. Results for these two experiments can be seen in rows 3-5 of
Figure \ref{fig:results}. Note that the photometric error depicted in the
rightmost column has been scaled up by a factor 1.5 for the sake of
visualization.

All results were implemented on a GPU for derivative computation, and the
optimization takes place on a CPU. Each optimization to converge took between
6-10 iterations of sequential Monte Carlo and gradient descent, which lasted
for approximately 10 minutes. Once the optimization had converged, rendering
the reference image took approximately 30 seconds.

\section{Discussion}
\label{sec:discussion}

The results from our two experiments using real world reference images clearly
show that the largest photometric errors often occur at depth discontinuities,
due to an imperfect reconstruction of the scene's geometry. This suggests that
we may improve the robustness of our algorithm by ignoring pixels near depth
discontinuities as performed \cite{ACM_2014_Zhou}.

We note that Figure \ref{fig:results} shows several departures from the
generative model that are present in the image results, which can be explained
through the natural behavior of our algorithm. The photometric error pane
illustrates that the curtain behind the scene in the first row experiences
greater photometric error as it recedes from the imaging plane. This is expected
as there is no difference in the environment map's imbuing of light in the
scene dependent on in-scene depth values (the light source is modeled at
infinite distance from the scene). Also, the latter three rows in
Figure \ref{fig:results} demonstrate that the parametrization of light using an
environment map do not cause a significant enough departure from an area light
that would cause the scene to not appear realistic. That said, there are
complicated patterns in the photometric error that result from this
parametrization which do not admit a simple resolution.

Finally and most importantly, our results demonstrate that our method may be
leveraged to both estimate light position and to realistically render scenes
with complicated optical phenomena, including diffusing and interacting shadows
and light sources. This approach is therefore a feasible option for both
augmented reality applications as well as in-scene parameter estimation.


\section{Conclusion}
\label{sec:conclusion}

We have presented a new algorithm for light source estimation in scenes
reconstructed using a RGB-D camera. Although we model light using an environment
map, we have shown that our algorithm can still accurately estimate lighting
conditions created by light sources located near the observed scene. Our major
contribution is developing a new technique that leverages the full expressive
power of the light transport equations to perform lighting optimization. The
presented optimization problem can potentially be reformulated to estimate any
term of the light transport equation, namely surface albedos, bidirectional
reflectance distribution functions, and scene geometry, providing a wealth of
directions for future research.


\bibliographystyle{kasper2017}
\bibliography{kasper2017}

\end{document}